\documentclass[sigconf]{acmart}
\AtBeginDocument{%
  }

\copyrightyear{2025}
\acmYear{2025}
\setcopyright{acmlicensed}\acmConference[SIGSPATIAL '25]{The 33rd ACM International Conference on Advances in Geographic Information Systems}{November 3--6, 2025}{Minneapolis, MN, USA}
\acmBooktitle{The 33rd ACM International Conference on Advances in Geographic Information Systems (SIGSPATIAL '25), November 3--6, 2025, Minneapolis, MN, USA}
\acmDOI{10.1145/3748636.3758021}
\acmISBN{979-8-4007-2086-4/2025/11}

\newif\ifdraft

\ifdraft
  \usepackage[colorinlistoftodos, textwidth=3.2cm, loadshadowlibrary, shadow]{todonotes}
\else
  \usepackage[disable]{todonotes}
\fi

\ifdraft
  \newcommand{\gilles}[1]{\todo[inline,color=blue!40,size=\scriptsize]{Gilles: #1}}
  \newcommand{\mahmoud}[1]{\todo[inline,color=orange!40,size=\scriptsize]{Mahmoud: #1}}
  \newcommand{\gaspard}[1]{\todo[inline,color=green!40,size=\scriptsize]{Gaspard: #1}}
\else
  \newcommand{\gilles}[1]{}
  \newcommand{\mahmoud}[1]{}
  \newcommand{\gaspard}[1]{}
\fi

\begin{document}

\title{Building a Foundation Model for Trajectory from Scratch}

\author{Gaspard Merten}
\email{gaspard.merten@ulb.be}
\orcid{ 0009-0007-7287-649X }
\affiliation{%
  \institution{Université libre de Bruxelles}
  \city{Brussels}
  \country{Belgium}
}

\author{Mahmoud Sakr}
\email{mahmoud.sakr@ulb.be}
\orcid{ 0000-0002-6741-8300 }
\affiliation{%
  \institution{Université libre de Bruxelles}
  \city{Brussels}
  \country{Belgium}
}

\author{Gilles Dejaegere}
\email{gilles.dejaegere@ulb.be}
\orcid{ 0000-0002-1526-9227 }
\affiliation{%
  \institution{Université libre de Bruxelles}
  \city{Brussels}
  \country{Belgium}
}

\renewcommand{\shortauthors}{Merten et al.}

\begin{abstract}
Foundation models are transformative in artificial intelligence, but building them from scratch, especially for mobility trajectories, is not yet clear or documented. This tutorial bridges this gap by demonstrating the steps and code of a minimal implementation of a trajectory-focused foundation model starting from GPT-2. Through a concise, step-by-step, code-driven process, we demonstrate adapting GPT-2 for spatiotemporal data. We then review and compare representative trajectory foundation models, such as TrajFM and TrajGPT, highlighting their architectural innovations and differences. Additionally, we introduce complementary techniques from related domains, like TimesFM's patching approach. Targeted at researchers and practitioners, this tutorial aims to explain the concepts and terminology of foundation models, at the implementation level. We find it timely and indispensable to create this educational material in order to support the SIGSPATIAL community in building and evaluating mobility foundation models, enhancing both research clarity and peer-review effectiveness in mobility AI.
\end{abstract}


\maketitle
Foundation models are a major emerging trend in artificial intelligence, with growing research interest across multiple domains, including mobility. Since the introduction of GPT, researchers have actively worked to replicate the success of large language models for other types of data, giving rise to the broader concept of foundation models. These models are trained on vast, diverse datasets to learn general patterns that can be adapted to many downstream tasks. In mobility, they can capture universal spatiotemporal behaviors from GPS traces, traffic flows, and transit data. Unlike specialized machine learning algorithms designed for single tasks, such as similarity calculation‑\cite{SIM17}, pattern matching~\cite{STP09}, or map-making~\cite{MM19}  foundation models offer a flexible, reusable base that can be fine-tuned for diverse applications.
~
The SIGSPATIAL community is progressively engaging with these developments, learning the concepts and the building blocks of foundation models, and their possible applications. However, many technical details remain vague unless one goes through hands-on implementation. While the research groups who are actively involved in foundation model development are competent to know these details, the rest of us (SIGSPATIAL community) might lack this level of knowledge, and we are still required to review these papers, and we further wish to adopt these methods in our research. 

This tutorial is designed with a dual goal: \textbf{to train the community on the hands-on implementation of foundation models} and \textbf{to make available an open source implementation of an \emph{educational} prototype for a trajectory foundation model}. As our conferences increasingly incorporate AI-related work, it becomes essential for the community to understand the terminology and concepts that underpin foundation models. By exposing how foundation models are built, step by step and at the code level, we hope to make their design and function more transparent. This will not only help participants develop a deeper understanding of the models themselves but also better equip them to review and evaluate related research contributions.

We do not present ourselves as experts in foundation models. Rather, we completed the exercise of building one ourselves and gained first-hand insight into its challenges and lessons. We believe that by sharing this experience, we can help others progress more quickly in their efforts, contributing to the broader advancement of research in this field.
~
~
To support these goals, the tutorial will introduce and clarify key technical terms and ideas through hands-on code examples. 



\section{Tutorial Outline}
The tutorial will be conducted iteratively. For each step, we will include a theoretical introduction to the different concepts, accompanied by a concrete implementation in Python code. This iterative approach follows a pedagogical structure that was inspired by the book \textit{Build an LLM From Scratch}~\cite{raschka2024build}, which offers a clear and practical flow for understanding foundation models.

Resources for learning how to build a mobility foundation model are still scarce. To address this gap, we begin with foundational concepts and incrementally build up the necessary components. Since many papers are explicitly inspired by Transformer and GPT architectures, we start with the architecture of Large Language Models (LLMs). The \textit{Build an LLM From Scratch} book provides an accessible entry point, offering tools and insights applicable to mobility foundation models. We first attempt a naive adaptation of GPT-2 for trajectory data, modifying each component, from the encoder and attention heads to the transformer blocks and decoder. Once this baseline model is functional, we refine the architecture to incorporate key techniques from the literature on mobility foundation models, such as ROPE, patching, and advanced masking. This step-by-step construction aims to build a granular and intuitive understanding of mobility foundation models, which we believe can benefit the broader community.

Based on this rationale, we propose structuring our tutorial as follows: We will first review the architecture of the GPT-2 model, then revise it to make it more suitable for trajectory data, drawing inspiration from relevant papers. Having built this intuitive comprehension, we will examine recent advancements in the field, comparing them to the educational model developed throughout the tutorial. Finally, we conclude with a summary of the tutorial and offer some insights on future directions.

\section{GPT-2 Architecture}

The first part of this tutorial will introduce the GPT-2 architecture~\cite{radford2019language}, which is directly derived from the Transformer architecture. GPT-2 is an LLM open-sourced by OpenAI, making it a suitable starting point for this tutorial. Moreover, it follows the decoder-only architecture, which is commonly used in mobility foundation models, and enables a transition from theoretical exposition to practical code implementation. Reimplementing GPT-2 from scratch offers valuable insights into its architecture and helps develop the intuition required to adapt it for mobility-related tasks.

Before exploring the GPT-2 architecture, we begin with a conceptual review of latent space, which plays a fundamental role in foundation models. A latent space encodes the model’s internal understanding of input data and serves as the basis for downstream tasks. To illustrate this concept, we use dynamic 3D visualizations that simplify the idea while retaining its core intuition.

We then transition to the architecture of GPT-2, starting with an explanation of tokenization, which plays a key role in LLMs. Although tokens are not always used in mobility foundation models, it remains relevant to review this foundational concept, particularly for audiences less familiar with AI.

From there, we demonstrate how tokenized inputs are transformed into vector representations using an embedding matrix, and explain the role of this matrix. Since embeddings do not capture positional information by themselves, we introduce the concept of positional encoding. This section includes both a visual walkthrough and a concise PyTorch implementation to demonstrate how a sentence is converted into a set of embeddings enriched with positional information.

Together, embeddings and positional encodings constitute the main input layers of the GPT-2 pipeline. The next major element is the Transformer block. We first focus on the attention mechanism, beginning with the problem it addresses, its mathematical formulation, and a simple code implementation. This discussion then expands to multi-head attention, explaining how it enables the model to capture multiple types of relationships in parallel.

Next, we present the full Transformer block, detailing its components, including the normalization layer, residual connections, feedforward layers, and the causal mask that ensures autoregressive behavior. This part is supported by code that implements the Transformer block as it appears in GPT-2.

Finally, we examine the output layer, where token probabilities are extracted from the output of the last Transformer block. This is particularly important, as words represent a limited, discrete set, quite different from the continuous coordinates encountered in mobility data, requiring later modifications.

By assembling these components, we reconstruct the overall \linebreak GPT-2 pipeline, consisting of a series of stacked Transformer blocks. We conclude this section with a brief overview of the training process, which will be discussed in more detail in the next part of the tutorial.

\section{Adapting GPT-2 to Trajectory Data}

In this second part, we present a concise end-to-end procedure for adapting the GPT-2 architecture to model trajectories, represented as sequences of $(\text{latitude}, \text{longitude}, t)$ triplets. The goal is not to build a production-grade mobility foundation model, but an educational one that highlights the key architectural adaptations required for spatiotemporal data.

Recent models such as TrajFM~\cite{lin_trajfm_2024}, TrajGPT~\cite{hsu_trajgpt_2024}, and TimesFM~\cite{lin_trajfm_2024} are all inspired by GPT-like architectures, even if this is not always explicitly stated. Rather than replicating these models, this tutorial simplifies the core concepts to provide a pedagogical bridge between general-purpose LLMs and domain-specific trajectory foundation models, helping to clarify their development.

The process begins with reworking the data pipeline to handle raw spatiotemporal sequences. Traditional tokenizers, designed for discrete, word-like inputs, are removed and replaced with custom parsers. Unlike natural language tokens, which map to a fixed vocabulary, trajectory data consists of continuous values such as latitude, longitude, and time that cannot be reliably tokenized. Instead, we adopt a preprocessing approach inspired by TrajFM, mapping each spatiotemporal point into a model-friendly structure with normalized coordinates and time features split into components such as day of week, hour, minute, and second. To ensure this encoding works as intended, we propose building a small pretext model, for instance, a simple autoencoder, to verify that the original positions can be reliably retrieved. The same autoencoder is reused once positional encoding is introduced to validate that the added transformations still preserve the underlying information. We found this approach of creating small test models particularly useful for debugging our methodology and confirming that each step behaves as expected.

Next, the GPT-2 embedding layer is replaced with a learnable projection layer that maps these inputs into a higher-dimensional space. This enables the model to encode richer information in subsequent transformer blocks. A concise PyTorch code example illustrates this custom embedding process. Following TrajGPT, we also adopt the original positional encoding from the "Attention is All You Need" paper rather than GPT-2’s standard approach.

Given the size of most mobility datasets, a streaming data loader is employed to efficiently feed the model without requiring all data to reside in memory. This ensures scalability and supports experimentation on commodity hardware. To facilitate rapid development and testing, we also introduce a synthetic dataset generator, which is particularly useful when real-world datasets are noisy or cumbersome to use.

To simplify training, prevent overfitting on smaller datasets, and align with recent studies~\cite{hsu_trajgpt_2024, lin_trajfm_2024}, we reduce the GPT-2 architecture to just two transformer blocks.

Instead of predicting absolute coordinates and timestamps, which can vary widely in scale, we use delta encoding. In this setup, the model predicts the changes in position and time between consecutive steps rather than the absolute values. This creates a more stable and learnable target. A minimal code snippet illustrates how this transformation is applied during preprocessing. Delta encoding also aligns naturally with autoregressive modeling, where each prediction depends on the previous step.

We further introduce a masked variant of the model, which allows it to fill in missing segments of a trajectory. This task generalizes next-step prediction because the model can still predict the next point while also handling gaps in the sequence.

The section concludes by demonstrating how to train the adapted model, compute evaluation metrics, and use the trained model in practice. We also highlight useful tools and provide references to freely available trajectory datasets for benchmarking and prototyping. The aim is for the attendees to follow the code during the tutorial, more information can be found in Section \ref{sec:materials}.

\section{State of the Art Trajectory Foundation Model}

Building on the hands-on walkthrough of the educational trajectory foundation model, this section contextualizes this implementation within the broader landscape of recent foundation models for mobility. It begins with an in-depth examination of TrajFM, a Transformer variant designed for trajectory modeling. Then it explores TrajGPT, which reframes trajectory prediction as a POI-sequence generation task. Finally, it surveys other notable approaches that extend Transformer-based architectures to spatiotemporal and time series domains, highlighting design patterns and innovations that inform the development of generalizable mobility foundation models.

\subsection{TrajFM: Enhancing Mobility Foundation Models}
The first paper we present is TrajFM, a mobility foundation model leveraging the Transformer block behind GPT-2. Rather than directly extending GPT-2, TrajFM innovatively alters the Transformer block to tailor it specifically for mobility scenarios. TrajFM, similar to our educational model, predicts continuous spatial coordinates and is designed explicitly for transferability across diverse geographic contexts. This section provides a structured exploration of TrajFM, emphasizing its distinctive innovations.

We organize our discussion around three primary contributions:

\begin{enumerate}
    \item \textbf{Advanced Encoding and Embedding Techniques}
    \item \textbf{Rotary Positional Embedding (RoPE)}
    \item \textbf{Sophisticated Masking Strategies}
\end{enumerate}

\paragraph{Advanced Encoding and Embedding Techniques.}
TrajFM incorporates both continuous coordinates (latitude and longitude) and discrete Points of Interest (POIs). To embed semantic information from POIs, it leverages OpenAI's embedding API, generating meaningful vectors based on textual descriptions. Spatial coordinates are normalized around the geographic center to enhance regional transferability and then projected into a higher-dimensional embedding space using a fully connected neural network, conceptually aligned with our earlier adaptation.

Temporal dimensions, such as day of the week, hour, minute, and elapsed trajectory time in seconds, are individually encoded using learnable Fourier encoding layers. These temporal embeddings are concatenated into a cohesive temporal representation, subsequently merged with spatial and POI embeddings, creating a comprehensive latent vector for each trajectory point.

\paragraph{Rotary Positional Embedding (RoPE) for Spatiotemporal Data.}\phantom{pl}
TrajFM notably employs Rotary positional embedding (RoPE)~\cite{su_roformer_2023}, specifically adapted for spatiotemporal data to enhance correlation modeling between trajectory points. RoPE leverages rotation matrices to capture both absolute and relative positional relationships effectively, thereby significantly improving long-range dependency modeling. We briefly illustrate how RoPE integrates within the attention mechanism, clarifying its implementation and benefits.

\paragraph{Sophisticated Masking Strategies.}
TrajFM uses two complementary masking techniques during training:

\begin{itemize}
    \item \textit{Dimension-specific masking}: selectively masking either temporal or spatial (coordinates/POI) dimensions at specific points.
    \item \textit{Segment masking}: masking entire trajectory segments.
\end{itemize}

Compared to our earlier approach, where we simply predicted the next token, TrajFM's masking strategy is notably more robust. It enables the model to support multiple downstream tasks, such as travel-time estimation, path inference, and trajectory completion. Segment masking, in particular, is directly inspired by Masked Language Modeling (MLM) strategies found in models such as BERT, where approximately 15\% of tokens are masked during training~\cite{devlin2018bert}.

\subsection{TrajGPT: Transformer-Based Region and Time Prediction}

While TrajFM predicts continuous coordinates, \textbf{TrajGPT} \cite{hsu_trajgpt_2024} specifically reframes trajectory prediction as a sequence prediction task of regions and associated temporal attributes, such as departure and arrival time. TrajGPT maintains a strong architectural lineage to the original Transformer code. Similarly to TimesFM, it cleverly leverages existing models tailored to the task, such as temporal embedding, instead of building everything from scratch. This is interesting to showcase to the audience and to uncover how one can compose existing architectures and adapt them into a newer model. 

This section highlights three essential architectural components of TrajGPT, demonstrating how transformer architectures can be adapted and specialized for mobility scenarios:

\begin{itemize}
    \item \textbf{Embedding Layer Stack}: TrajGPT replaces token embeddings with Time2Vec \cite{kazemi2019time2vec} and Space2Vec \cite{mai2020multispace2vec} encoders, combined with a learnable region embedding. This modular design enables effective handling of continuous spatiotemporal data, providing flexibility and task agnosticism. We briefly outline minimal PyTorch implementations to demonstrate the simplicity and utility of these embeddings.

    \item \textbf{Region-Level Prediction}: Rather than directly outputting coordinates, TrajGPT predicts discrete regions or Points of Interest (POIs), employing categorical loss functions and softmax layers. This approach simplifies training and aligns conceptually with token prediction in language models, facilitating a direct comparison to our previous continuous-coordinate modeling and enabling us to discuss the trade-offs between these methodologies.

    \item \textbf{Multi-head Output Modules}: TrajGPT uses specialized prediction heads, such as Gaussian Mixture Models (GMM), to estimate travel times and visit durations. This design choice effectively captures the multimodal distribution of temporal patterns observed in mobility data, offering superior accuracy compared to simpler regression methods. We discuss the conceptual advantages of these modules and illustrate their applicability to complex temporal dynamics.
\end{itemize}

In this section, we demonstrate how TrajGPT builds upon transformer architectures, extending them into domain-specific modules tailored to spatiotemporal mobility tasks, and contrast these advancements with our earlier educational model implementation.

\subsection{Other Noteworthy Approaches}

To conclude this tutorial, we briefly highlight several additional innovative approaches to modeling trajectories and time series. These selected examples illustrate alternative techniques that complement or combine existing methods, broadening the scope of mobility research.

\textit{TimesFM}~\cite{das_decoder-only_2024} is a foundation model developed by Google, specifically designed for large-scale time series data. Building on \linebreak Transformer-based architectures similar to GPT, TimesFM introduces the concept of \emph{patching}, where consecutive values are grouped into single embedding vectors, significantly reducing the sequence length and computational overhead. This idea originally appeared in the influential paper \textit{``A Time Series is Worth 64 Words''}~\cite{nie2022time}.

\textit{TrajGDM}~\cite{chu_trajgdm_2023} provides a distinct approach by integrating Generative Adversarial Networks (GANs) with Transformer blocks. This hybrid structure enables the model to generate realistic synthetic trajectories that accurately mirror the underlying distribution of mobility data, making it especially valuable for data augmentation and privacy-preserving applications.


Through this concise overview, participants gain insight into the diverse methods and ideas currently shaping trajectory modeling, encouraging further exploration and innovation within the field.

\section{Materials}
\label{sec:materials}
All tutorial materials are available in our open-source GitHub repository (\url{https://github.com/GaspardMerten/CheminTF}). They provide clear, concise PyTorch\footnote{\url{https://pytorch.org/}} code for each step, a modular implementation of the trajectory foundation model, data loaders for large-scale mobility datasets, and training and evaluation scripts. The repository also includes a notebook folder with a Jupyter notebook for following the presentation, as well as multiple notebooks retracing the steps we took to build the foundation model, with some steps left for the user to complete, offering an exercise-style hands-on experience. A list of available datasets for model training is also provided. In this tutorial, we build on the open-source code of GPT-2\footnote{\url{https://github.com/openai/gpt-2}} in the first part, and leverage TrajGPT\footnote{\url{https://github.com/ktxlh/TrajGPT}} source code in the third part, ensuring participants gain practical experience with state-of-the-art trajectory modeling frameworks.

Participants are welcome to explore the code before the session and even install it, but there is no expectation to write code during the session, as we will focus on presenting code snippets. For motivated participants, however, the presentation can be followed interactively in a Google Colab notebook, with a link for attendees to create their own instance provided in the README.md of the tutorial repository.

\begin{acks}
This work was partially funded by the EU’s Horizon Europe research and innovation program under Grant No. 101070279 MobiSpaces, and  EMERALDS under Grant No. 101093051.
\end{acks}

\bibliographystyle{ACM-Reference-Format}
\bibliography{sample-base}

\end{document}